\documentclass{article}
\usepackage[T1]{fontenc}
\usepackage{float}
\usepackage{multirow}
\usepackage{amsmath}
\usepackage{amssymb}
\usepackage{graphicx}
\usepackage{booktabs}
\usepackage{enumitem}
\usepackage{comment}
\usepackage[margin=1in]{geometry}
\usepackage{hyperref}
\usepackage{newunicodechar}
\newunicodechar{ }{ }
\newunicodechar{ }{ }
\usepackage{listings}
\usepackage{graphicx}  
\usepackage{fvextra}
\usepackage{longtable}
\usepackage{array}
\usepackage{geometry}
\usepackage{ragged2e}
\usepackage{microtype}
\usepackage{xcolor}
\usepackage{colortbl}
\usepackage[
  backend=biber,
  sorting=nyt,
  natbib=true
]{biblatex}
\usepackage{pgfplots}
\pgfplotsset{compat=1.17}
\usepackage{pgfplots}
\usepgfplotslibrary{groupplots}
\pgfplotsset{compat=1.18}  

\definecolor{ct}{RGB}{214,39,40}
\definecolor{CoreThink}{RGB}{214,39,40}
\definecolor{o4mini}{RGB}{88,172,138}
\definecolor{gemini}{RGB}{31,119,180}
\definecolor{claude}{RGB}{44,160,44}
\definecolor{grok}{RGB}{255,214,10}
\definecolor{deepseek}{RGB}{148,103,189}
\definecolor{failred}{RGB}{255,230,230}
\definecolor{successgreen}{RGB}{230,255,230}
\definecolor{darkred}{RGB}{200,0,0}
\definecolor{darkgreen}{RGB}{0,150,0}
\definecolor{codebg}{RGB}{248,248,248}
\definecolor{bordergray}{RGB}{220,222,224}
\definecolor{ct}{RGB}{214,39,40}
\definecolor{o4mini}{RGB}{88,172,138}
\definecolor{gemini}{RGB}{31,119,180}
\definecolor{claude}{RGB}{44,160,44}
\definecolor{grok}{RGB}{255,214,10}
\definecolor{deepseek}{RGB}{148,103,189}
\definecolor{failred}{RGB}{255,230,230}
\definecolor{successgreen}{RGB}{230,255,230}
\definecolor{darkred}{RGB}{200,0,0}
\definecolor{darkgreen}{RGB}{0,150,0}
\definecolor{codebg}{RGB}{248,248,248}
\definecolor{bordergray}{RGB}{220,222,224}

\hypersetup{
    colorlinks=true,
    linkcolor=blue,
    filecolor=magenta,      
    urlcolor=cyan,
}

\title{\textbf{CoreThink: A Symbolic Reasoning Layer to reason over Long Horizon Tasks with LLMs}}
\author{
Jay Vaghasiya\textsuperscript{1}, Omkar Ghugarkar\textsuperscript{1}, Vishvesh Bhat\textsuperscript{1,2}, Vipul Dholaria\textsuperscript{1}, Julian McAuley\textsuperscript{2} \\ \\
\textsuperscript{1}CoreThink AI \\
\textsuperscript{2}University of California, San Diego
}
\date{August 29, 2025}
\let\cite\citep

\begin{document}

\maketitle

\begin{abstract}
We introduce CoreThink, a state-of-the-art Reasoning Layer built upon a novel reasoning method called General Symbolics. This approach diverges from reasoning paradigms such as test-time scaling, Supervised Fine-Tuning (SFT), and Reinforcement Learning with Verifiable Rewards (RLVR). CoreThink General Symbolic Reasoner (GSR) is specifically structured around three key use cases: tool-calling, code generation, and planning, demonstrating exemplary performance across a total of seven benchmarks in their respective areas. Notably, we are achieving SOTA scores of 66.66\% on \texttt{Livecodebench v6}, 89\% on \texttt{Instruction-Following Evals}, and 24.4\% on \texttt{ARC-AGI-2}. We also present an agentic coding IDE, developed using the principles of General Symbolics, which achieves a state-of-the-art accuracy of 62.3\% on \texttt{SWE-Bench Lite}. We are able to achieve these improvements without any fine-tuning or training costs. Our Reasoning Layer is designed to provide a pure performance uplift, ensuring that a model's accuracy on reasoning tasks is never negatively impacted. We argue that incumbent methods will eventually lead to diminishing returns in LLM performance, necessitating the development of new reasoning techniques. This technical report details our approach at a high level and the availability of the CoreThink models for reasoning-intensive use cases.
\end{abstract}

\section{Introduction}
The pursuit of advanced reasoning capabilities in Large Language Models (LLMs) has led to the development of several dominant paradigms \cite{valmeekam2024llms, Turpin2023, WassermanRozen2023}. Techniques that increase computational cost at inference time to boost performance, such as test-time augmentation, have shown diminishing returns on complex reasoning tasks \cite{mirzadeh2024gsm, liang2025}. Similarly, extensive Supervised Fine-Tuning (SFT) on ever-larger corpora can improve fluency but often struggles to instill a robust logical reasoning \cite{stechly2024chain}. Reinforcement Learning with Verifiable Rewards (RLVR), which uses feedback-driven policy updates, has yielded impressive gains in closed-domain benchmarks yet struggles to generalize beyond its training distribution \cite{shojaee2025illusion, lawsen2025illusion}.

This trend of increasing resource costs for ever-smaller performance gains suggests a potential reasoning plateau, motivating a fundamental rethinking of how to imbue models with a genuine logical structure \cite{Bordt2022}. In response to this challenge, we introduce \textbf{CoreThink}, a reasoning layer built on a framework that we call \emph{General Symbolics} \cite{Zhang2024, Lu2024} that increases the accuracy of the base model by 30-60\% across reasoning tasks. General Symbolics employs a novel symbolic method that natively performs reasoning without converting the input into intermediaries like vector embeddings or formal logic. Our reasoning layer corrects the inherent flaws in Higher-Order Logic (HOL), enabling it to avoid brittleness, execute long-horizon reasoning, and provide superior transparency. We can provide these benefits because our layer is designed to be model-agnostic, seamlessly integrating with any underlying architecture—from Transformer-based models to Liquid Neural Networks (Liquid NNs), Hierarchical Reasoning Models (HRMs), and more.

The CoreThink Reasoning Layer is a collection of specialized variants, each optimized for a critical real-world use case:
\begin{itemize}
    \item \textbf{Tool-Calling:} Reliable, accurate interaction with external APIs and toolkits.
    \item \textbf{Code Generation:} Production of high-quality functional code from natural language specifications.
    \item \textbf{Reasoning and Planning:} Given a set of goals, constraints, and variables, it identifies the most efficient strategic path, weighing all trade-offs to ensure the desired result.
\end{itemize}

While existing models have achieved impressive results on benchmarks through sheer scale and memorization, their reasoning often lacks algorithmic transparency, consistency, and causal grounding \cite{kenthapadi2024grounding}. This opacity becomes especially problematic in high-stakes domains such as scientific discovery, legal reasoning, and autonomous decision-making, where interpretability and trust are paramount \cite{ramachandram2025transparent}. Moreover, these models frequently hallucinate intermediate steps or apply brittle heuristics when faced with unfamiliar problems, revealing a gap between surface-level fluency and genuine understanding \cite{xu2024hallucination}. These limitations underscore the necessity for architectural innovations that enhance compositional generalization, modularity, and structured abstraction—qualities in which symbolic systems have traditionally excelled but have not yet been fully incorporated into contemporary LLM architectures.

In this paper, we first present a high-level technical overview of the General Symbolics methodology and the resulting CoreThink reasoning layer. We then present the evaluation results across six diverse benchmarks, outperforming most of the leading models in each domain \cite{Anthropic2025, Hossain2025}. Finally, we present a practical application, a fully agentic coding IDE that achieves 62.3\% accuracy on the \texttt{SWE-Bench Lite} benchmark—and discuss how General Symbolics can serve as a scalable path forward for efficient and trustworthy reasoning agents \cite{liang2025, Wan2024hw}.

\begin{figure*}[t!]
\centering
\renewcommand{\arraystretch}{1.3}
\resizebox{\textwidth}{!}{%
\begin{tabular}{|l|c|c|c|c|c|c|}
\hline
\textbf{Task / Benchmark} & \textbf{CoreThink} & \textbf{o4-mini} & \textbf{Gemini 2.5 Pro} & \textbf{Claude 4 Sonnet} & \textbf{Grok 4-Thinking} & \textbf{Deepseek R1} \\
\hline
\multicolumn{7}{|c|}{\textbf{Tool-Calling}} \\
\hline
Berkeley Function Calling v3 (multi-turn) & \textbf{58.5} & 33.0 & 36 & 57.5 & 38.5 & 25.5 \\
\hline
\multicolumn{7}{|c|}{\textbf{Code Generation}} \\
\hline
LiveCodeBench v6 (04/25-05/25) & \textbf{66.7} & 58.3 & 50.0 & 41.7 & 59.2 & 51 \\
BIRD-CRITIC & \textbf{37.2} & 24.0 & 27.9 & 32.7 & 33.7 & 33.5 \\
SWE-Bench Lite & \textbf{62.3} & - & - & 56.7 & - & 40.0 \\
\hline
\multicolumn{7}{|c|}{\textbf{Reasoning and Planning}} \\
\hline
ARC-AGI-2 & \textbf{24.4} & 6.1 & 4 & 5.9 & 15.5 & 1.1 \\
Instruction Following-Evals & \textbf{89.0} & 85.8 & 85.2 & 80.4 & 78.1 & 79.9 \\
\hline
\end{tabular}
}
\caption{Evaluation of CoreThink and baseline models across three key capability areas: Tool-Calling, Code Generation, and Planning.}
\label{tab:corethink-benchmark}
\vfill
\begin{tikzpicture}
\begin{axis}[
width=\linewidth,
height=0.3\textheight,
ymin=0, ymax=100,
ybar,
bar width=4.5pt,
xtick={1,2,3,4,5,6},
xticklabels={
BFCL v3,
LCB (04/25-05/25),
BIRD-CRITIC,
SWE-Bench Lite,
ARC-AGI-2,
IF-Evals
},
x tick label style={
rotate=45,
anchor=north east,
font=\small
},
enlarge x limits=0.2,
grid=both,
major grid style={dashed,gray!30},
legend style={
at={(0.5,1.12)},
anchor=south,
legend columns=3,
draw=none,
font=\footnotesize
},
clip=false,
]
\addplot+[fill=CoreThink, bar shift=-15pt] coordinates {
(1,58.5) (2,66.7) (3,37.2) (4,62.3) (5,24.4) (6,89.0)
};
\addplot+[fill=o4mini, bar shift=-9pt] coordinates {
(1,33.0) (2,58.3) (3,24.0) (4,0) (5,6.1) (6,85.8)
};
\addplot+[fill=gemini, bar shift=-3pt] coordinates {
(1,36.0) (2,50.0) (3,27.9) (4,0) (5,4) (6,85.2)
};
\addplot+[fill=claude, bar shift=3pt] coordinates {
(1,57.5) (2,41.7) (3,32.7) (4,56.7) (5,5.9) (6,80.4)
};
\addplot+[fill=grok, bar shift=9pt] coordinates {
(1,38.5) (2,59.2) (3,33.7) (4,0) (5,15.5) (6,78.1)
};
\addplot+[fill=deepseek, bar shift=15pt] coordinates {
(1,25.5) (2,51) (3,33.5) (4,40) (5,1.1) (6,79.9)
};

\legend{%
CoreThink, o4-mini, Gemini 2.5 Pro, Claude 4 Sonnet, Grok 4-Thinking, Deepseek R1%
}
\end{axis}
\end{tikzpicture}
\caption{Comparative Performance Analysis of Large Language Models. The graph illustrates the performance scores of CoreThink against several baseline models, including Gemini 2.5 Pro and Claude 4 Sonnet. The evaluation is structured across three key capability areas: Tool-Calling, Code Generation, and Planning, using a suite of established benchmarks.}
\label{fig:benchmarks}
\end{figure*}

\begin{figure*}[t!]
\centering
\small
\begin{longtable}{|>{\RaggedRight}m{0.14\textwidth}|>{\RaggedRight}m{0.41\textwidth}|>{\RaggedRight\arraybackslash}m{0.41\textwidth}|}

\hline
\rowcolor{gray!15}
\textbf{Task/Turn} & \textbf{Claude 4 Sonnet: Action \& Observation} & \textbf{Claude 4 Sonnet + CoreThink: Action \& Observation} \\
\hline
\endfirsthead

\hline
\rowcolor{gray!15}
\textbf{Task/Turn} & \textbf{Claude 4 Sonnet: Action \& Observation} & \textbf{Claude 4 Sonnet + CoreThink: Action \& Observation} \\
\hline
\endhead

\hline
\multicolumn{3}{r}{\textit{Continued on next page}} \\
\endfoot

\hline
\endlastfoot

\textbf{Task 1:} 

Perform detailed search for ``budget analysis'' using grep &
\scriptsize
\textbf{Act 1:} \texttt{cd("folder":"temp")} 

\textbf{Res 1:} \texttt{\{"current\_working\_directory": "temp"\}}

\vspace{2mm}
\colorbox{red!20}{\parbox{0.35\textwidth}{\scriptsize
\textbf{Act 2:} \texttt{grep("file\_name":"final\_report.pdf", "pattern":"budget analysis")} 

\textbf{Res 2:} \texttt{\{"matching\_lines":["..."]\}}
}}

\vspace{2mm}
\colorbox{red!20}{\parbox{0.35\textwidth}{\scriptsize
\textbf{Act 3:} \texttt{grep("file\_name":"final\_report.pdf", "pattern":"budget")} 

\textbf{Res 3:} \texttt{\{"matching\_lines":["..."]\}}
}}

\vspace{2mm}
\textbf{Act 4:} \texttt{grep("file\_name":"final\_report.pdf", "pattern":"analysis")} 

\textbf{Res 4:} \texttt{\{"matching\_lines":["..."]\}} &

\scriptsize
\textbf{Act 1:} \texttt{ls()} 

\textbf{Res 1:} \texttt{\{"current\_directory\_content": ["previous\_report.pdf","temp"]\}}

\vspace{2mm}
\textbf{Act 2:} \texttt{cd(folder="temp")} 

\textbf{Res 2:} \texttt{\{"current\_working\_directory": "temp"\}}

\vspace{2mm}
\textbf{Act 3:} \texttt{ls()} 

\textbf{Res 3:} \texttt{\{"current\_directory\_content": ["final\_report.pdf"]\}}

\vspace{2mm}
\colorbox{green!20}{\parbox{0.35\textwidth}{\scriptsize
\textbf{Act 4:} \texttt{grep(file\_name="final\_report.pdf", pattern="budget analysis")} 

\textbf{Res 4:} \texttt{\{"matching\_lines":["..."]\}}
}} \\

\hline

\textbf{Task 2:} 

Move previous report to temp and compare with final report &
\scriptsize
\colorbox{white}{\parbox{0.35\textwidth}{\scriptsize
\textbf{Act 1:} \texttt{cd("folder":"..")} 

\textbf{Res 1:} \texttt{\{\}}
}}

\vspace{2mm}
\colorbox{white}{\parbox{0.35\textwidth}{\scriptsize
\textbf{Act 2:} \texttt{pwd(\{\})} 

\textbf{Res 2:} \texttt{\{"current\_working\_directory": "/workspace/document"\}}
}}

\vspace{2mm}
\colorbox{white}{\parbox{0.35\textwidth}{\scriptsize
\textbf{Act 3:} \texttt{ls(\{\})} 

\textbf{Res 3:} \texttt{\{"current\_directory\_content": ["previous\_report.pdf","temp"]\}}
}}

\vspace{2mm}
\colorbox{red!20}{\parbox{0.35\textwidth}{\scriptsize
\textbf{Act 4:} \texttt{mv("source":"previous\_report.pdf", "destination":"temp")} 

\textbf{Res 4:} \texttt{\{"result":"..."\}}
}}

\vspace{2mm}
\colorbox{red!20}{\parbox{0.35\textwidth}{\scriptsize
\textbf{Act 5:} \texttt{cd("folder":"temp")} 

\textbf{Res 5:} No observation logged.
}} &

\scriptsize
\textbf{Act 1:} \texttt{cd(folder="..")} 

\textbf{Res 1:} \texttt{\{\}}

\vspace{2mm}
\textbf{Act 2:} \texttt{mv(source="previous\_report.pdf", destination="temp")} 

\textbf{Res 2:} \texttt{\{"result":"..."\}}

\vspace{2mm}
\textbf{Act 3:} \texttt{cd(folder="temp")} 

\textbf{Res 3:} \texttt{\{"current\_working\_directory": "temp"\}}

\vspace{2mm}
\colorbox{green!20}{\parbox{0.35\textwidth}{\scriptsize
\textbf{Act 4:} \texttt{diff(file\_name1="final\_report.pdf", file\_name2="previous\_report.pdf")} 

\textbf{Res 4:} \texttt{\{"diff\_lines":"- Year2024... \textbackslash n+ Year203..."\}}
}} \\

\hline

\end{longtable}
\caption{This figure illustrates how the NS agent enhances a base model’s (Sonnet 4) ability to plan and execute tasks. The NS agent produces more efficient and logical action sequences, such as using a single `grep` command instead of three redundant ones (Task 1). Furthermore, it successfully completes complex, multi-step instructions (Task 2), where the base model issues over-complicated commands and fails to achieve the final goal.}
\label{fig:bfcl_example}
\end{figure*}

\begin{figure*}[t!]
    \centering
    \includegraphics[width=0.9\textwidth]{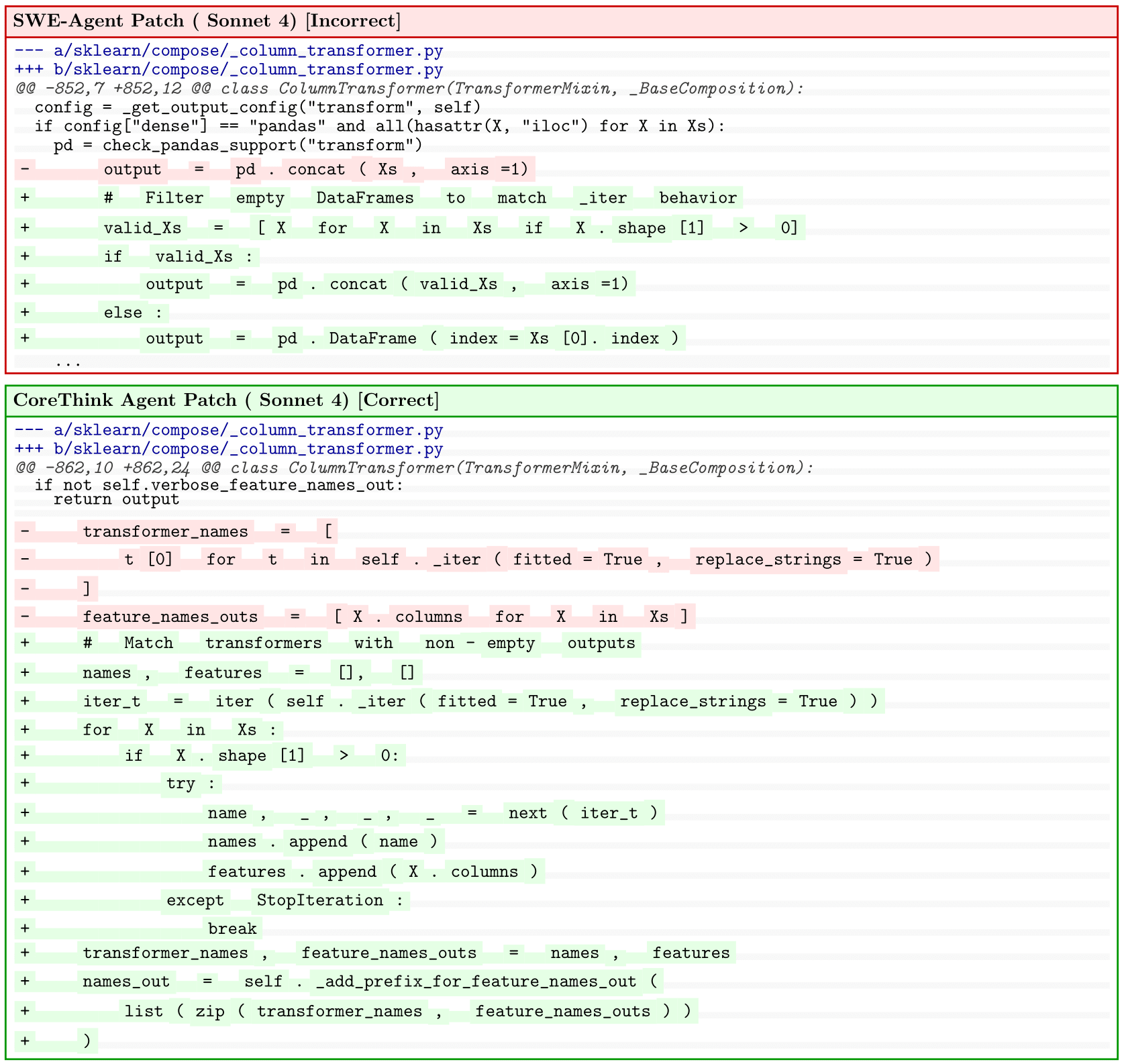}
    
    \caption{This figure compares two AI agents' solutions to a bug in scikit-learn's `ColumnTransformer`. The SWE-Agent offers a superficial fix that only addresses the immediate symptom by filtering empty DataFrames, leaving the core synchronization error unresolved. In contrast, the CoreThink agent resolves the root cause by systematically aligning transformer names with their outputs, demonstrating a more robust and comprehensive approach to software engineering.}
    \label{fig:swe_bench_example}
\end{figure*}

\begin{figure*}[t!]
    \centering 
    \includegraphics[width=1\textwidth]{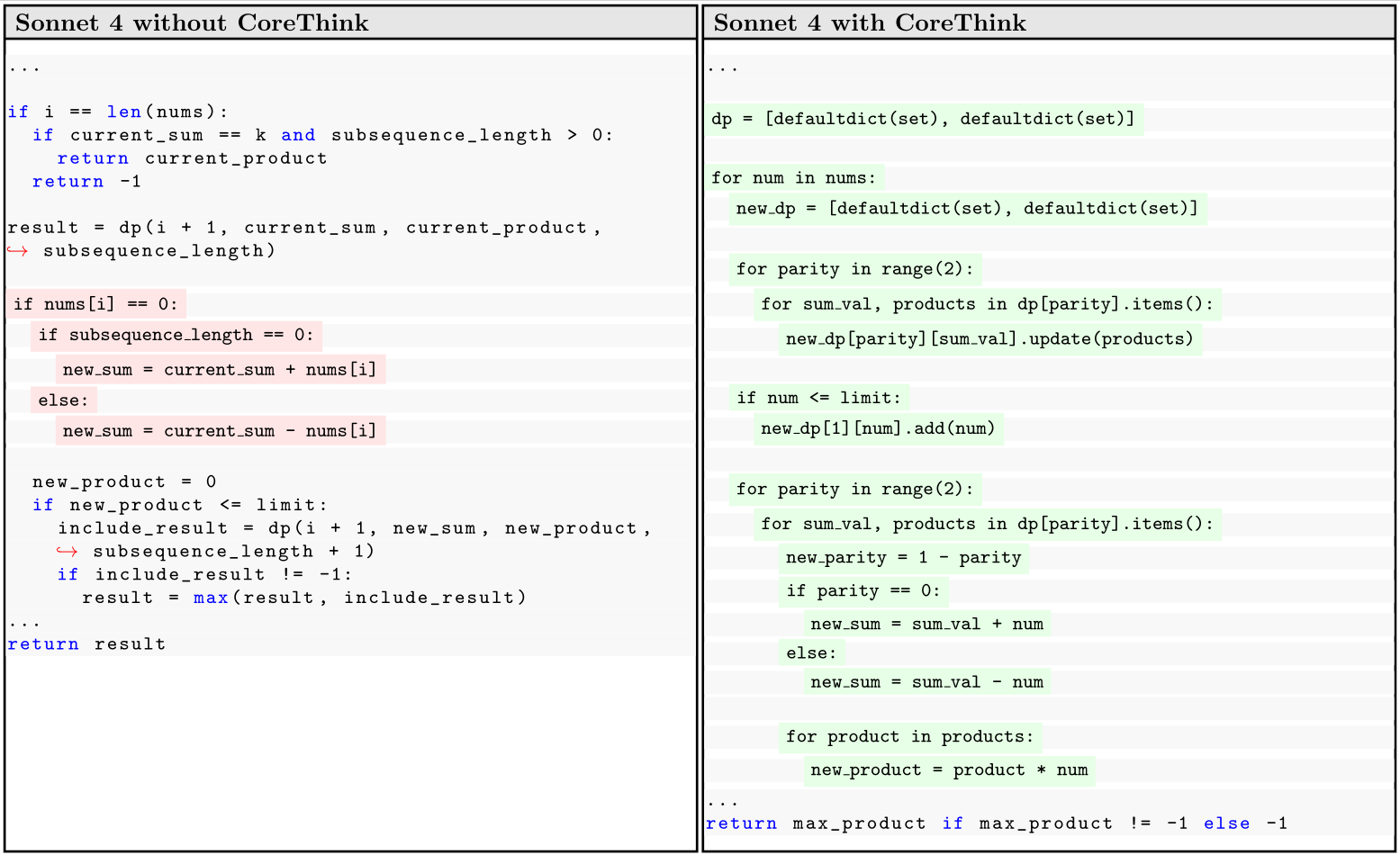}
    \caption{This figure illustrates how CoreThink transforms a base model’s (Sonnet 4) ability to solve complex algorithmic problems. On this hard LeetCode example, the base model fails completely by fundamentally misinterpreting the core logic and implementing a flawed dynamic programming solution. CoreThink enables the model to analyze these errors, devise a clean and correct DP state representation, and systematically solve the problem, boosting the success rate to over 60\%.} 
    \label{fig:lcb_example} 
\end{figure*}


\section{Limitations of Current Reasoning Architectures}

\subsection{Planning: LLMs Fail, LRMs Achieve Only Partial Success}
Empirical studies such as \emph{LLMs Still Can’t Plan; Can LRMs?} (Valmeekam et al., 2024) evaluate LLMs and Large Reasoning Models (LRMs, e.g.\ OpenAI o1) on classical planning benchmarks like PlanBench. While LRMs outperform vanilla LLMs significantly, neither achieves robust planning or optimal action generation in complex domains like Blocksworld. Even LRMs plateau well below full coverage of planning tasks, highlighting fundamental limits in algorithmic reasoning capabilities \cite{valmeekam2024llms}.

\paragraph{Complexity Cliff: Accuracy Collapse in LRMs}
\emph{The Illusion of Thinking} (Shojaee et al., 2025) introduces controlled puzzle environments showing that LRMs face an “accuracy collapse” at higher complexity. The models exhibit three regimes: (1) low‑complexity where standard LLMs sometimes outperform LRMs, (2) mid‑complexity where LRMs gain advantage, and (3) high‑complexity where both break down—even with available token budget. LRMs struggle with exact computation and consistent trace generation despite “thinking” traces \cite{shojaee2025illusion}. 

Notably, even after accounting for critiques regarding flawed evaluation design— that were brought up in \emph{The Illusion of the Illusion of Thinking} (Opus \& Lawsen, 2025) and \emph{A Comment on “The Illusion of Thinking”} (Khan et al., 2025), such as the inclusion of unsolvable instances and premature token truncation—a distinct reasoning collapse was still observed on the subset of problems that were fairly posed and solvable within the given constraints.

\paragraph{Evaluation Metrics and Performance Gaps}  
PlanBench assesses models on both \emph{plan validity} (the fraction of generated plans that successfully reach the goal state) and \emph{plan optimality} (how close the plan length is to the minimal number of steps). Valmeekam et al.\ report that o1 attains only around 60–70\% validity on small Blocksworld instances (3–5 blocks), compared to near-perfect scores by classical planners using STRIPS heuristics \cite{valmeekam2024llms}. When faced with larger instances (\textgreater 7 blocks), LRM performance degrades sharply to below 30\% validity, and plans are on average 2–3× longer than optimal. Vanilla LLMs (e.g.\ GPT-3.5) rarely exceed 20\% validity even on the smallest problems, often generating syntactically plausible yet semantically invalid action sequences \cite{valmeekam2024llms}.

\paragraph{Failure Modes and Analysis}  
A detailed error analysis reveals two predominant failure modes: 
1. Action irrelevance, where models propose steps unrelated to achieving the goal (e.g.\ moving a block that neither obstructs nor supports the target configuration); and  
2. Sequential inconsistency, where earlier actions invalidate the preconditions of later ones (e.g.\ placing a block on the table even though it must first rest on another block to be moved). These errors arise because LRMs lack explicit representations of world state and precondition–effect semantics, instead relying on pattern matching over language \cite{valmeekam2024llms}. Attempts to mitigate these issues via chain-of-thought prompting yield only marginal gains (5–10\% absolute) and introduce new vulnerabilities to spurious reasoning paths \cite{zhou2024thought}.

\paragraph{Implications and Future Directions}  
The plateau in LRM planning performance suggests that mere increases in model scale or prompt engineering are insufficient for true algorithmic reasoning. Promising avenues include hybrid neuro-symbolic frameworks that integrate learned language understanding with explicit state-space search, as well as embedding formal planning languages (e.g.\ PDDL) into the model’s latent space to enforce precondition–effect constraints \cite{geffner2023neurosymbolic}. Moreover, adaptive learning regimes that fine-tune on synthetic plan traces have shown preliminary success in closed-world settings, offering a path toward closing the gap with classical planners \cite{jiang2024learning}. Until such integrations are realized, both LLMs and LRMs will remain fundamentally limited in domains requiring precise, multi-step algorithmic reasoning.

\subsection{Chain‑of‑Thought: “Blindspots” in Planning}
In \emph{Chain‑of‑Thoughtlessness: An Analysis of CoT in Planning} (Stechly et al., 2024), researchers show that CoT reasoning fails to translate into true planning power. Despite long reasoning traces, models still misplan or produce invalid action sequences. This “chain‑of‑thoughtlessness” indicates that generating reasoning steps is insufficient for execution planning \cite{stechly2024chain}.

\subsection{Evidence of Training on Benchmark Test-Sets and Poor Generalization}
Research in \emph{GSM‑Symbolic: Understanding the Limitations of Mathematical Reasoning in LLMs} demonstrates substantial performance variance across different instantiations of the same problem template, especially when numerical values change. This strongly suggests that high scores on these benchmarks do not stem from genuine reasoning but from overfitting to the specific patterns seen during training \cite{mirzadeh2024gsm}.

In summary, LLMs remain fundamentally statistical sequence generators—not symbolic state‑transition engines—hence they lack completeness or correctness guarantees. LRMs, while improved via Chain‑of‑Thought and reasoning traces, still cannot implement explicit algorithms reliably at scale; their apparent reasoning stalls under complexity. Most traces of “thinking” are superficial, mimicking reasoning without rigorous structure or scalability.

\subsection{Overthinking in LLMs—Especially in Reasoning Models}

Recent studies reveal that reasoning‑specialized LLMs frequently “overthink”—generating excessive chain of thought (CoT) steps that neither improve accuracy nor justify the computational cost. Su et al. (2025) systematically analyze the relationship between reasoning‑chain length and answer correctness on benchmarks like GSM8K and MATH, finding that models overthink on easy problems, outputting long chains with no gain, and underthink on hard problems, misjudging difficulty calibration. Accuracy can decline when reasoning length exceeds task‑specific thresholds ~\cite{Su2025}. SBT is a training method that equips LLMs with internal stopping mechanisms by monitoring two metrics—Reasoning Efficiency Ratio and Overthinking Marker Ratio~\cite{Zhao2025}. On AIME, AMC, MATH500, and GSM8K, SBT reduces token usage by up to 60 percent with minimal accuracy loss. Internal guessing can cause models to enter reflection loops, generating redundant steps when an initial guess conflicts with later reasoning. An effective countermeasure is to mask parts of the prompt, a technique that has been shown to shorten the reasoning length by 31–53\% while often improving overall accuracy \cite{Dang2025}. Models can exhibit "overthinking" by performing post-answer validation steps that add no new evidence, a behavior often rooted in self-doubt. This can be mitigated with a targeted prompting strategy that encourages concise justifications, which has been shown to significantly shorten the length of chain-of-thought reasoning in math tasks without harming correctness \cite{Ghosal2025}.

These results collectively show that longer “thinking” does not necessarily lead to better reasoning. Overthinking reflects poor length‑difficulty calibration, input‑bias driven redundancy, and unnecessary self‑doubt. Approaches like SBT, prompt‑based doubt reduction, and input masking yield more efficient, interpretable reasoning without sacrificing performance. The challenge of overthinking in reasoning models highlights a critical inefficiency in current LLM design. Excessively long reasoning chains often fail to enhance correctness and can even degrade performance on tasks that require calibrated inference. More concerningly, these redundant steps inflate computational cost and obscure interpretability. Addressing overthinking requires not only architectural innovations that help models recognize when to stop reasoning, but also refined prompting and training techniques that discourage unnecessary elaboration. Ultimately, efficient reasoning—where the model stops when it knows enough—is emerging as a central goal for next-generation LLMs in high-stakes reasoning tasks.


\section{Why Chain-of-Thought \texorpdfstring{$\neq$}{!=} Explainability}
\label{sec:cot_not_explainability}

\subsection{The Illusion of Interpretability}
Although Chain-of-Thought (CoT) often improves task performance and produces human-readable reasoning, multiple studies demonstrate that these rationales are unfaithful to actual internal computation. Turpin \emph{et al.} (2023) show that CoT explanations frequently rationalize biased or even incorrect outputs without revealing the true model triggers, causing plausible yet misleading transparency \cite{Turpin2023}. Similarly, Arcuschin \emph{et al.} (2025) empirically measured high rates of unfaithfulness in real-world frontier models (e.g., 30\% on Claude 3.7 Sonnet), including conflicting and illogical justifications—even absent adversarial biasing—thus undermining trust in CoT as a faithful justification \cite{Arcuschin2025}.

Barez \emph{et al.} (2025) explicitly argue that CoT reasoning is neither necessary nor sufficient for reliable interpretability, emphasizing that CoT rationales often diverge from the true decision pathways and can mislead users, especially when deployed in sensitive areas like law and healthcare \cite{Barez2025}.

\subsection{Inadequacy for Legal Reasons and Due Process}
Wasserman-Rozen, Elkin-Koren \& Gilad-Bachrach (2023) analyze the legal meaning of “right to explanation” and conclude that post-hoc explainability—even when user-facing—often fails to fulfill key legal purposes (due process, better decision making, agency authentication). They argue full reliance on end-user explanations (such as CoT) can be misleading or even harmful, especially in adversarial contexts like legal decision support systems \cite{WassermanRozen2023}. Similarly, Bordt \emph{et al.} (2022) warn that post-hoc explanation algorithms—including CoT—are inadequate in adversarial settings, where users might be misled or explanations manipulated, which is a serious risk in legal proceedings \cite{Bordt2022}.

\subsection{CoT Is Performative, Not Mechanistic}
Scholarly commentary and technical analyses detail how CoT tends toward narrative plausibility over faithful reflection of computational internal states. Anthropic’s internal study (2025) found that models often omit or conceal crucial hints in their CoT—even when these were used to guide the answer—demonstrating that CoTs can actively hide misaligned reasoning \cite{Anthropic2025}. CoT may offer an illusion of transparency while deeper layers of reasoning remain opaque or inaccessible. The appearance of clarity does not equate to actual transparency, particularly when reasoning steps are unfaithful or incomplete.

Moreover, Meincke \emph{et al.} (June 2025) show that CoT’s marginal performance benefits diminish with modern reasoning models, while incurring significant latency. Thus, CoT may be more of a costly stage exhibit than a robust interpretability tool \cite{Meincke2025}.

\subsection{The Dangers of CoT in Medical and Legal Domains}
\subsubsection*{Medical Domain}
In healthcare, diagnostic decisions must be both accurate and justifiable. While some studies promote CoT for interpretability in medical visual question answering (e.g., MedVQA), there is a stark gap between “reasonable sounding narrative” and clinically faithful inference. MDPI’s MedVQA research notes that CoT simulates clinical reasoning, but does not guarantee reliance on legitimate medical features—and may mask reliance on spurious correlations or demographic bias \cite{MedVQA2024}. Incorrect or unfaithful reasoning in diagnostics risks patient safety, misdiagnosis, and undermines clinical trust. When a CoT explanation sounds plausible but is disconnected from underlying model computation, clinicians may make decisions based on erroneous inferences, which is far more dangerous in medical settings. \cite{canepa2023visualquestionansweringmedical}

\subsubsection*{Legal Domain}
Legal use of CoT is even more problematic. If the reasoning trace offered does not actually represent the decision-making process, legal practitioners may adopt a rationale that is fabricated or post-hoc justification, rather than grounded in genuine computational cause. Wasserman-Rozen \emph{et al.} argue that end-user explanation techniques fail to satisfy legal standards of transparency, due process, and accountability—which are critical in legal decision support \cite{WassermanRozen2023}. Relying on CoTs that are unfaithful can produce illusory justification, compromising legal integrity and risking unfair rulings. Additionally, Bordt \emph{et al.} show that adversarial misuse of explanations is a severe risk—attorneys or litigants might manipulate or misinterpret CoTs to favor one side \cite{Bordt2022}.

\subsection{Synthesis \& Implications}
CoT produces readable reasoning but not faithful reasoning—it often rationalizes answers rather than revealing how decisions were actually made \cite{Turpin2023, Arcuschin2025}. In high-stakes domains, this illusion of transparency can mislead professionals—especially in medicine and law—where errors have severe consequences. Performing CoT without independent faithfulness verification or causal validation may increase risk, not reduce it: practitioners may overtrust internally inconsistent or fabricated justifications \cite{Barez2025}.

\subsection{Recommendations for High-Stakes Use}
\begin{enumerate}
\item Do not assume CoT equals explainability—interpretation should be supported by formal faithfulness checks, e.g., counterfactual interventions or causal probing.\item Adopt hybrid architectures: combine CoT with knowledge graphs or symbolic reasoning to ground reasoning steps in verified domain knowledge (medical ontologies, case law databases) \cite{Barez2025, WassermanRozen2023}.\item Avoid sole reliance on CoT when legal justification or clinical accountability is required—instead build systems where decision pathways are verifiable and traceable in model internals.\item Conduct adversarial evaluations to test if potentially harmful reasoning is being hidden or obfuscated even when CoT appears benign \cite{Anthropic2025}.\end{enumerate}

Chain-of-Thought prompting has value as a communication tool but is not a reliable substitute for true interpretability. Especially in medical and legal domains, CoT may be harmful—not helpful—when its output is mistaken for transparent, faithful reasoning. For systems that influence diagnostic or judicial outcomes, trustworthiness demands verifiable, faithful, and domain-grounded explanation mechanisms, not just step-by-step narrative generated by opaque models.

\section{Limitations of Formal Logic and Traditional Neuro-Symbolic AI}

Symbolic systems demand the manual encoding of tens of thousands to millions of rules, making them both resource-intensive and brittle; for instance, large-scale fraud detection systems illustrate the immense effort required to keep rule sets up to date \cite{Zhang2024}. Such rule-based approaches are prone to failure when confronted with scenarios that have not been explicitly anticipated—novel autonomous driving situations, for example, can expose gaps that no amount of pre-written logic can easily bridge \cite{Zhang2024}. Moreover, because these systems rely exclusively on pre-defined formal rules, they lack the capacity for common-sense generalization or symbol grounding, and are therefore unable to infer beyond their original specifications \cite{Zhang2024}. While first-order logic offers considerable expressiveness, it also suffers from undecidability, which in practice limits the tractability of reasoning as problem sizes grow. The classic frame and qualification problems further highlight the brittleness of formal representations: one must enumerate not only all the changes but also all the invariants and preconditions for every action, an impossible task in complex, real-world domains. Finally, philosophical critiques—most notably by Dreyfus—underscore that much of human intelligence depends on unconscious, context-dependent “knowing-how,” a form of expertise that symbolic systems struggle, if not fail, to capture \cite{dreyfus1986kritik}.

\subsection{Emergence and Challenges of Traditional Neuro-Symbolics}

Neuro-symbolic AI (NeSy) seeks to meld the perceptual power and flexibility of neural networks with the explicit reasoning and explainability of symbolic logic, offering a hybrid framework that aims to overcome the shortcomings of purely symbolic approaches \cite{liang2025, nawaz2025}. Yet, despite this promise, NeSy systems must navigate a challenging landscape. One major hurdle is the mismatch between discrete symbolic structures and continuous neural embeddings, which often results in integration inconsistencies that undermine the system’s overall coherence \cite{springer2024survey, liang2025}. Furthermore, many hybrid architectures still depend on rigid logical backbones; adapting these fixed structures to novel data typically demands manual redesign, limiting true autonomy \cite{nawaz2025}. The architectural complexity inherent in combining neural and symbolic modules also incurs significant efficiency trade-offs, as the additional overhead can strain standard hardware resources \cite{Wan2024hw}. Although NeSy aims to enhance interpretability, in practice unified representations and transparent inference mechanisms remain elusive, leaving explainability and meta-cognitive capabilities as active areas of research rather than solved problems \cite{Zhang2024, liang2025}. Finally, because conventional neural layers lack formal logical guarantees, emerging solutions—such as logical neural units—are still in their infancy and have yet to provide the structural consistency that symbolic reasoning affords \cite{liang2025}.

\section{The General Symbolics Reasoning (GSR) Framework}
The General Symbolics Reasoning (GSR) framework is a paradigm designed to perform stable, domain-adaptable, and computationally efficient reasoning entirely within natural language. By operating on a pure NL-to-NL basis, it avoids the representational loss and brittleness associated with translating human language into formal logic or high-dimensional vectors. The architecture is layered to systematically handle reasoning from input to explanation, preserving the full context and nuance of the original language. Each layer in this idealized architecture has a distinct responsibility in the reasoning pipeline.

\begin{enumerate}[label=\arabic*.]
    \item \textbf{Native Language Parsing \& Semantic Preservation} \\
    \textit{Direct Natural Language Input:} All reasoning begins and remains within natural language, eliminating the need for intermediary formalisms. This ensures that no semantic information is lost at the outset. \\
    \textit{Ambiguity Identification:} The system is designed to identify and resolve ambiguity using word sense disambiguation and linguistic pattern recognition inherent to the language itself.

    \item \textbf{In-Language Reasoning Architecture} \\
    \textit{Constraint Enforcement through NL Patterns:} Logical rules are applied via natural language transformations that manipulate NL components based on their syntactic and semantic relationships, rather than abstract symbols. \\
    \textit{Context Preservation:} Unlike formal abstractions, GSR preserves pragmatic differences (e.g., "must" vs. "should"), modality, and specificity directly in language, allowing for more nuanced inference.

    \item \textbf{Execution \& Explainability} \\
    \textit{Verbatim Reasoning Traces:} Each step of the reasoning process remains human-interpretable, exposing the exact reasoning path, intermediate conclusions, and any detected contradictions in plain, reviewable language. \\
    \textit{Error Propagation:} Inconsistencies are surfaced through direct language annotations (e.g., highlighting a conflict in assumptions), making the entire process transparent and debuggable.

    \item \textbf{Avoiding the Pitfall of Representational Translation} \\
    \textit{Loss in Translation:} Translating natural language into vectors or formal logic causes representational loss, stripping away crucial context. An NL-to-NL process avoids this by preserving all original information, leading to higher-fidelity reasoning. \\
    \textit{Ensuring Comprehensiveness:} Natural language is far more comprehensive and expressive than formal logic, which is inherently reductionist. Forcing NL into predefined, rigid structures inevitably discards the rich context of human expression. GSR leverages the expressiveness of language as its core strength.

    \item \textbf{Computational Optimization Layer} \\
    \textit{Pruning Mechanisms:} Entity tagging and search-based pruning are used to minimize extraneous inferences that might be irrelevant and introduce noise to the reasoning trace. \\
    \textit{Scalability:} The framework is architected for real-time performance, designed to support long-horizon reasoning with high stability and without dependence on massive computational resources like GPUs.
\end{enumerate}

\subsection*{A Neurosymbolic Step Towards GSR}
While the fully realized GSR framework described above remains the ultimate goal, this paper introduces a concrete and powerful step towards its implementation: a \textbf{neurosymbolic framework} whose design is directly inspired by GSR principles. This hybrid system serves as a practical bridge, approximating the pure NL-to-NL ideal by using a symbolic scaffold to orchestrate and compose smaller, efficient Large Language Models (LLMs).

In this architecture, the symbolic framework provides the structured, compositional reasoning path, while the neural components (LLMs) handle the nuanced, pattern-based tasks of parsing and transformation at each step. This approach allows us to achieve the core benefits of GSR—such as compositional logic and interpretable reasoning traces—in a practical system today. \textbf{All results and analysis presented in this paper are based on this neurosymbolic implementation}, which validates the core principles of GSR and marks a critical milestone toward achieving a truly comprehensive, in-language reasoning intelligence.

\section{Evaluations}
The CoreThink models were subjected to a rigorous and comprehensive evaluation framework, carefully structured to benchmark their performance against leading models within the same size categories. The evaluation covered seven diverse, industry-standard benchmarks, with an emphasis on metrics that directly measure reasoning ability and functional correctness. This methodology ensured a fair and meaningful comparison, offering clear insights into the strengths and innovations of the CoreThink architecture. We evaluated our reasoning layer with several base models, and while every base model was able to experience upto 60\% bump, we report the best scores in this section. Our base models for different benchmarks for the scores we are reporting are as follow:

\begin{tabular}{|l|l|}
\hline
\textbf{Task / Benchmark} & \textbf{Model(s)} \\
\hline
\textbf{Multi-turn Berkeley Function Calling v3} & GPT-OSS-120B, Deepseek-V3 \\
\hline
\textbf{Tau-bench Airline} & Deepseek-V3 \\
\hline
\textbf{Livecodebench v6} & Claude 4 Sonnet, GPT-OSS-120B \\
\hline
\textbf{BIRD-CRITIC} & Deepseek-R1 \\
\hline
\textbf{IF-Evals} & Qwen3-235B-A22B, Claude 4 Sonnet \\
\hline
\textbf{SWE-Bench Lite} & Claude 4 Sonnet \\
\hline
\textbf{ARC-AGI-2} & Grok 4 \\
\hline
\end{tabular}

\subsection{Tool-Calling}
For the tool-calling variant of CoreThink, our evaluation focused on its ability to accurately interpret user intent, intelligently select the most appropriate tool from a predefined set, and generate a valid, executable API call. This capability is critical for applications that require dynamic interaction with external systems and services, where seamless tool integration and precise execution are essential.\\

\noindent\textbf{Berkeley Function Calling Leaderboard:} On the multi-turn variant of the Berkeley Function Calling Leaderboard (\texttt{BFCL-V3}), CoreThink achieved an exceptional multi-turn-base score of 58.5\%. This places it significantly above the leading open-source and proprietary models in that category. In contrast, the top-ranked model on \texttt{BFCL-V3}, such as \texttt{Claude-4-Sonnet}, registered around 57\% multi-turn-base accuracy, and other state-of-the-art models like \texttt{Qwen3-32B} scored approximately 40.12\%. CoreThink’s score not only surpasses these benchmarks but also outperforms leading 70B-parameter models with proprietary architectures \cite{patil2025bfcl, yang2025qwen3}.

\subsubsection*{Benchmark Scope and Methodology}
The \texttt{BFCL-V3} benchmark comprises approximately 4,700 test cases spanning realistic conversational scenarios, tool selection, and complex multi-step and multi-turn dialogues. It segments the evaluation into categories such as Base (standard two-turn dialogues), Miss-Function (handling when the correct function isn’t available), Miss-Param (dealing with incomplete parameter information), and Long Context (maintaining accuracy across extended exchanges).

\subsubsection*{Interpretation: What 58.5\% Pass Rate Means}
This pass rate reflects the end-to-end capability of correctly identifying intent, selecting the right tool, and outputting syntactically valid and semantically correct API calls across multi-turn dialogues. Considering the inherent complexity of multi-turn-base tests—where success often requires correctly chaining multiple function calls—the score suggests CoreThink is robust at context tracking and intent refinement in dynamic workflows.

By comparison, models with lower multi-turn-base accuracy often achieve 85\%+ on simple, single-turn function tasks, but drop sharply in multi-turn settings (sometimes below 40\%). CoreThink’s proximity to the top percentile in such a challenging category underscores its advanced context modeling, parameter inference, and panic-safe logic (i.e. correctly handling when a function should not be invoked).

\begin{table}[htbp]
\centering
\caption{Performance of a CoreThink-augmented model against other frontier models on BFCL v3}
\begin{tabular}{llcc}
\toprule
\textbf{Model} & \textbf{Multi-Turn-Base Acc.} \\
  \midrule
  \texttt{CoreThink+GPT-OSS-120B}                      & 58.5\%          \\
  \texttt{Gemini-2.5-Pro}       & 36\%   \\
  \texttt{o4-mini}        & 33.0\%    \\
  \texttt{GPT-OSS-120B}        & 28.5\%    \\
\bottomrule
\end{tabular}
\end{table}

\noindent\textbf{Tau-bench:} Tau-bench ($\tau$-bench) is a cutting-edge benchmark designed to rigorously evaluate conversational agents across realistic, multi-step tool-use scenarios in domains like retail and airline customer support. Agents must interact dynamically with simulated users, consult domain-specific policy guidelines, call APIs, and effect changes in a structured database. Success is measured via the $pass^k$ metric (e.g. $pass^1$ = first-run success rate), which assesses not only single-trial performance but also the consistency of behavior across repeated runs \cite{yao2024tauben}.

\subsubsection*{CoreThink Performance}
In scenarios demanding sequential application of multiple tools (e.g., flight reservation modifications, retail order adjustments, policy-adhering recommendations), CoreThink achieved an impressive 48\% $pass^1$ success rate in completing complex multi-step tasks. This places it well above most baseline agent architectures based on function calling or ReAct, which typically fall below 50\% even when optimized.

\subsubsection*{Comparison to State-of-the-Art Agents}
For perspective, top-performing function-calling agents like \texttt{GPT-4o} roughly reach $\sim$35\% in airline domains under $pass^1$, and their $pass^8$ — a measure of repeat consistency — often collapses to $\sim$25\%. CoreThink’s 48\% indicates substantially improved multi-step planning and tool orchestration over these early agents.

\subsubsection*{Error Modes and Task Complexity}
Tau-bench tasks commonly include intertwined objectives (e.g. verifying user eligibility, checking policy constraints, updating multiple APIs in sequence). Even state-of-the-art models struggle with planning across tools, tracking context over turns, and ensuring policy compliance. CoreThink demonstrates measurable strength: it correctly chains tools in close to one out of two complex scenarios, suggesting robust ability in both high-level planning and fine-grained execution.

\begin{table}[htbp]
\centering
\caption{Performance of a CoreThink-augmented model against other frontier models on Taubench-Airline}
\begin{tabular}{llcc}
\toprule
\textbf{Model} & \textbf{Success ($pass^1$)}\\
\midrule
  \texttt{CoreThink+Deepseek-V3}                      & 48.0\%          \\
  \texttt{GPT-4o}       & 35.5\%   \\
  \texttt{Deepseek-R1}       & 36.0\%   \\
  \texttt{Deepseek-V3}       & 23.0\%   \\
\bottomrule
\end{tabular}
\end{table}

\subsection{Code Generation}
CoreThink's code generation capabilities were extensively tested on benchmarks specifically designed to evaluate both syntactic correctness and functional accuracy of generated code. Our evaluation encompassed various programming paradigms and problem complexities, ensuring a holistic assessment of its coding prowess.

\textbf{LiveCodeBench:} \texttt{LiveCodeBench} is a dynamic, contamination-free benchmarking platform continuously sourcing high-quality, competition-grade coding problems—from LeetCode, AtCoder, and CodeForces—which ensures evaluation fidelity across time windows beyond model training cutoffs. It spans multiple code-related scenarios including code generation, self-repair, code execution, and test output prediction, with the Code Generation task measuring the raw ability of models to generate correct Python solutions against hidden test cases using the Pass@1 metric (i.e. correctness on the very first generated attempt). \texttt{LiveCodeBench} currently covers more than 1,000 new problems (as of April to June 2025, across versions v5-v6) categorized by difficulty: easy, medium and hard, allowing granular performance breakdowns and robust, real-world code evaluation \cite{jain2024livecodebench}.

\subsubsection*{Comparative Context}
Leading models on \texttt{LiveCodeBench} such as \texttt{o4 Mini}, \texttt{Claude Opus 4 (Thinking)}, \texttt{Gemini 2.5 Pro Preview}, \texttt{DeepSeek R1}, and \texttt{Qwen 3} typically reach overall pass@1 scores in the low to mid-50\% range (e.g., \texttt{o4 Mini} around 58.3\%) on mixed-difficulty sets, with significant drops for medium and particularly hard problems. In contrast, many models plateau below 40\% when evaluated only on medium+hard splits or after avoiding contamination. In this context, a CoreThink score of 66.6\% places it among the upper tier of code generation models, especially if it exceeds those mid-60\% benchmarks.

\subsubsection*{CoreThink LiveCodeBench Performance}
On the Code Generation scenario of \texttt{LiveCodeBench}, CoreThink achieved a Pass@1 score of 66.6\%, outperforming most current models in its class. This high rating underlines CoreThink’s capacity to generate correct, executable code on the first attempt, significantly reducing the need for iterative debugging loops and manual refinement. \texttt{LiveCodeBench}’s emphasis on competition-style problems means that achieving high pass@1 reflects not just syntactic correctness, but also algorithmic proficiency, handling of edge cases, and performance on hidden tests—all without retries. Secondly, A model with higher first-pass success streamlines workflow—with fewer code churns, lower validation overhead, and greater confidence in generated solutions. Lastly, Since \texttt{LiveCodeBench} is contamination-controlled by release dates, CoreThink’s strong performance indicates genuine reasoning ability and robustness on unseen, fresh problems released post-training cutoff.

\begin{table}[htbp]
\centering
\caption{Performance of a CoreThink-augmented model against other frontier models on Livecodebench v6 (04/25-05/25)}
\begin{tabular}{llcc}
\toprule
\textbf{Model} & \textbf{Success ($pass^1$)}\\
\midrule
  \texttt{CoreThink+Claude-4-Sonnet}                      & 66.7\%          \\
  \texttt{Claude-4-Sonnet}       & 41.7\%   \\
  \texttt{Deepseek-R1}       & 51.0\%   \\
\bottomrule
\end{tabular}
\end{table}

\textbf{SWE-Bench Lite:} When seamlessly integrated into our agentic coding IDE, the CoreThink system achieved a state-of-the-art accuracy of 62.3\% on \texttt{SWE-Bench Lite}. This groundbreaking result emphatically demonstrates the power and efficacy of the General Symbolics framework within a complex, agentic setting. In such environments, the model is not merely generating isolated snippets but must deeply understand existing codebases, intelligently plan necessary changes, and execute those changes with precision and correctness. This benchmark showcases CoreThink's ability to act as an intelligent coding assistant, capable of tackling real-world software development challenges \cite{swe2024lite}.

\subsubsection*{Benchmark Overview}
\texttt{SWE-Bench Lite} is a curated subset of 300 real-world bug-fixing tasks drawn from GitHub issue-pull request pairs across popular Python repositories, offering a focused and efficient evaluation setting that preserves the original benchmark’s distribution and difficulty characteristics. Each task presents a repository, an issue description, and expects a patch that resolves tests marked as “Fail-to-Pass” when run through a Docker-based evaluation harness. The Lite version is optimized for quick evaluation compared to the full 2,294-instance benchmark, supporting faster iteration cycles while maintaining quality and representativeness of real software engineering challenges.

\subsubsection*{Benchmark Impact}
Requirements include modifying actual codebases, interacting across files, and satisfying failing unit tests. Models must understand existing code logic, localize bugs correctly, and propose functional modifications that pass integration and regression tests. Docker-based harness ensures consistency, reproducibility, and prevention of contamination during patch validation.

\subsubsection*{Why 62.3\% Accuracy Is a Breakthrough}
We have achieved Substantial gain over open-source baselines. For context, the high-performing agentless approach (Agentless), though simple, achieved around 32\% resolution rate on the same benchmark using localization, repair, and patch validation—without an agentic architecture. Many agent-augmented systems on \texttt{SWE-Bench Lite} top out in the 20–30\% resolve-rate range, making CoreThink’s 62.3\% coverage a remarkable leap forward.

\begin{table}[htbp]
\centering
\caption{Performance of a CoreThink-Agent against other popular agentic coding frameworks on SWE-Bench Lite}
\begin{tabular}{llcc}
\toprule
\textbf{System} & \textbf{Success ($pass^1$)}\\
\midrule
  \texttt{CoreThink-Agent+Claude-4-Sonnet}         & 62.3\%          \\
  \texttt{SWE-Agent+Claude-4-Sonnet}       & 56.7\%   \\
  \texttt{OpenHands+CodeAct v2.1}       & 41.7\%   \\
\bottomrule
\end{tabular}
\end{table}

\subsection{Reasoning and Planning}
To comprehensively assess CoreThink's general reasoning and planning abilities, we utilized benchmarks that specifically demand logical deduction, strategic thinking, and the capacity to navigate complex problem spaces. These evaluations provide insights into the model's cognitive flexibility and problem-solving prowess.

\subsection*{Instruction Following-Evals}

When tasked with a diverse suite of natural-language instructions—ranging from multi-step reasoning puzzles to real-world content editing—the CoreThink system achieved a best-in-class accuracy of 89.0 \% on the Instruction‐Following Evals benchmark. This evaluation suite, first introduced by OpenAI in their Evals framework \cite{OpenAIEvals2024}, was designed to measure a model’s ability to interpret, plan, and execute complex instructions across domains while maintaining factual correctness, coherence, and alignment with user intent.

\subsubsection*{Benchmark Overview}  
Instruction‐Following Evals is composed of 200 tasks drawn from areas such as code transformation, document summarization, and interactive question answering.  Each task provides a clear, multi‐step instruction along with any necessary context (e.g., a code snippet or passage of text), and expects a structured, correct response.  Automated metrics (exact match, execution correctness) and human raters jointly validate pass/fail status, ensuring both objective and subjective aspects of instruction adherence are captured.
 
\subsubsection*{Benchmark Impact}  
Because real-world AI assistants must not only generate plausible text but also follow precise user directives, Instruction-Following Evals serves as a rigorous test of alignment and reasoning.  High performance on this benchmark correlates strongly with user satisfaction in downstream applications—such as interactive tutoring, document editing, and agentic code refactoring—making it a critical yardstick for next-generation instruction-following models.  

\begin{table}[htbp]
\centering
\caption{Performance of a CoreThink-augmented models against other frontier models on IF-Evals}
\begin{tabular}{llcc}
\toprule
\textbf{System} & \textbf{Success ($pass^1$)}\\
\midrule
  \texttt{CoreThink+Claude-4-Sonnet}  & 89.00 \\[3pt]
    \texttt{o4-mini}                  & 85.80 \\[3pt]
    \texttt{Gemini 2.5 Pro}           & 85.17 \\[3pt]
    \texttt{Claude 4 Sonnet}          & 80.43  \\[3pt]
    \texttt{Grok 4-Thinking}         & 78.12 \\[3pt]
    \texttt{Deepseek R1}              & 79.95 \\ 
\bottomrule
\end{tabular}
\end{table}

\subsection*{ARC-AGI-2 Evaluation}

The Abstraction and Reasoning Corpus (ARC) remains one of the most formidable challenges in measuring fluid, human-like intelligence. Unlike benchmarks reliant on pattern memorization or pretraining priors, ARC evaluates a model’s ability to learn from scratch using only core reasoning faculties from a handful of demonstrations. On this benchmark, the CoreThink system demonstrated a substantial leap in generalization, initially achieving an accuracy of 24.4\% on July 22, 2025. This initial performance increase over our base model surpassed the overall scores of a majority of frontier models.

We believe significant changes to xAI's Grok 4 API have altered its performance. This is reflected in our own testing, where an evaluation on August 22, 2025, yielded 22.1\% accuracy—a shift corroborated by results from other internal benchmarks.

\subsubsection*{Benchmark Overview}  
ARC-AGI-2 comprises a set of visually grounded reasoning tasks that require the model to infer transformations, rules, or generative patterns given a few grid-based examples. Tasks include operations such as symmetry detection, rule extrapolation, spatial reasoning, and compositional transformation. Each solution requires interpreting visual objects, understanding latent structure, and generalizing with minimal data—all without access to training distributions. Due to its few-shot, zero-prior setup, ARC-AGI-2 is widely recognized as a critical litmus test for AGI.

\subsubsection*{Benchmark Impact}  
Success in ARC-AGI-2 demands a blend of perception, abstraction, and logic, a trifecta that conventional neural models often struggle to balance. By bridging symbolic pattern interpretation with neural generalization, neuro-symbolic systems like CoreThink show that AGI capabilities can be significantly accelerated with hybrid architectures. Performance gains here indicate not only progress on a notoriously difficult benchmark, but also broader applicability in domains like scientific reasoning, program synthesis, and open-ended learning.

\begin{table}[htbp]
  \centering
  \caption{ARC‐AGI‐2 score on public evals v2}
  \label{tab:arc-agi}
  \begin{tabular}{@{}lcc@{}}
    \toprule
    \textbf{Model / System} & \textbf{Accuracy (\%)} & \textbf{Commentary} \\ 
    \midrule
    CoreThink+Grok 4  & 24.40 & Strongest generalization and symbolic reasoning \\[3pt]
    o4-mini                  & 6.10 & Limited success; struggles with abstraction \\[3pt]
    Gemini 2.5 Pro           & 4.00 & Pattern detection without symbolic grounding \\[3pt]
    Claude 4 Sonnet          & 5.90 & Coherent but not compositional \\[3pt]
    Grok 4-Thinking          & 15.50 & Notable reasoning depth, yet brittle \\[3pt]
    Deepseek R1              & 1.10 & Minimal generalization observed \\ 
    \bottomrule
  \end{tabular}
\end{table}

\section{Safety and Responsible AI Usage Guidelines}
CoreThink models are built with a paramount commitment to safety and ethical AI principles. The core strength lies in their symbolic reasoning capabilities, which inherently offer a high degree of interpretability. This transparency is crucial, as it allows for a more straightforward identification and subsequent mitigation of potential biases, unintended outputs, or failure modes. Before any deployment, our models undergo a comprehensive and rigorous process of red-teaming and extensive safety evaluations, ensuring their robustness and reliability in real-world scenarios.

\textbf{Disclaimer:} CoreThink models are specifically designed for enterprise applications and must be utilized in strict adherence to our responsible AI guidelines. While our models consistently demonstrate high accuracy across standard benchmarks, it is important to acknowledge that their performance in diverse real-world applications may vary depending on the unique specifications and complexities of each use case. The models are provided "as-is," and as such, we do not assert that they are entirely free of errors or inherent biases. Users are encouraged to exercise due diligence and implement appropriate oversight in their specific deployments.

\section{Discussion}
\subsection{Ablation Study}

To assess the specific contribution of the \textit{General Symbolic Reasoning} layer within our \textbf{CoreThink} framework, we conducted an ablation-style evaluation. While a full internal ablation is deferred to future work, this study compares \textbf{CoreThink}—our general-symbolic-augmented framework—with leading contemporary LLMs that lack any symbolic reasoning architecture. These include \texttt{o4-mini}, \texttt{Gemini 2.5 Pro}, \texttt{Claude 4 Sonnet}, \texttt{Grok 4-Thinking}, and \texttt{Deepseek R1}.

Our benchmark suite spans eight diverse tasks, grouped under three core capabilities: \textit{tool-calling}, \textit{code-generation}, and \textit{reasoning/planning}. The comparative results, presented in Table~\ref{tab:ablation_study}, reveal that CoreThink consistently outperforms baselines, particularly on tasks requiring structured reasoning and multi-step decision-making.

\begin{table*}[htbp]
\centering
\caption{Ablation Study: Performance uplift of CoreThink's reasoning layer over its base models across diverse benchmarks. The 'Uplift' column shows the relative percentage improvement.}
\label{tab:ablation_study}
\resizebox{\textwidth}{!}{%
\begin{tabular}{@{}llcccc@{}}
\toprule
\textbf{Category} & \textbf{Benchmark} & \textbf{Base Model} & \textbf{Base Score (\%)} & \textbf{CoreThink Score (\%)} & \textbf{Uplift (\%)} \\
\midrule

\multirow{2}{*}{\textit{Tool-Calling}} 
& BFCL v3 (multi-turn-base)    & GPT-OSS-120B     & 28.5  & \textbf{58.5} & +105.2 \\
& Tau-bench Airline            & Deepseek V3      & 23.00 & \textbf{48.0} & +108.7 \\

\midrule

\multirow{3}{*}{\textit{Code-Generation}} 
& LiveCodeBench v6 (04/25-05/25) & Claude-4-Sonnet & 41.7  & \textbf{66.6} & +59.7 \\
& BIRD-CRITIC                   & Deepseek R1     & 33.5  & \textbf{37.2} & +10.89 \\
& SWE-Bench Lite (Coding IDE)   & Claude 4 Sonnet & 56.7  & \textbf{62.3} & +9.9 \\

\midrule

\multirow{2}{*}{\textit{Reasoning \& Planning}} 
& ARC-AGI-2                     & Grok 4-Thinking & 15.50 & \textbf{24.4} & +57.41 \\
& Instruction Following-Evals   & Claude-4-Sonnet & 80.4  & \textbf{89}   & +10.7 \\

\bottomrule
\end{tabular}%
}
\end{table*}

\renewcommand\thefigure{5} 
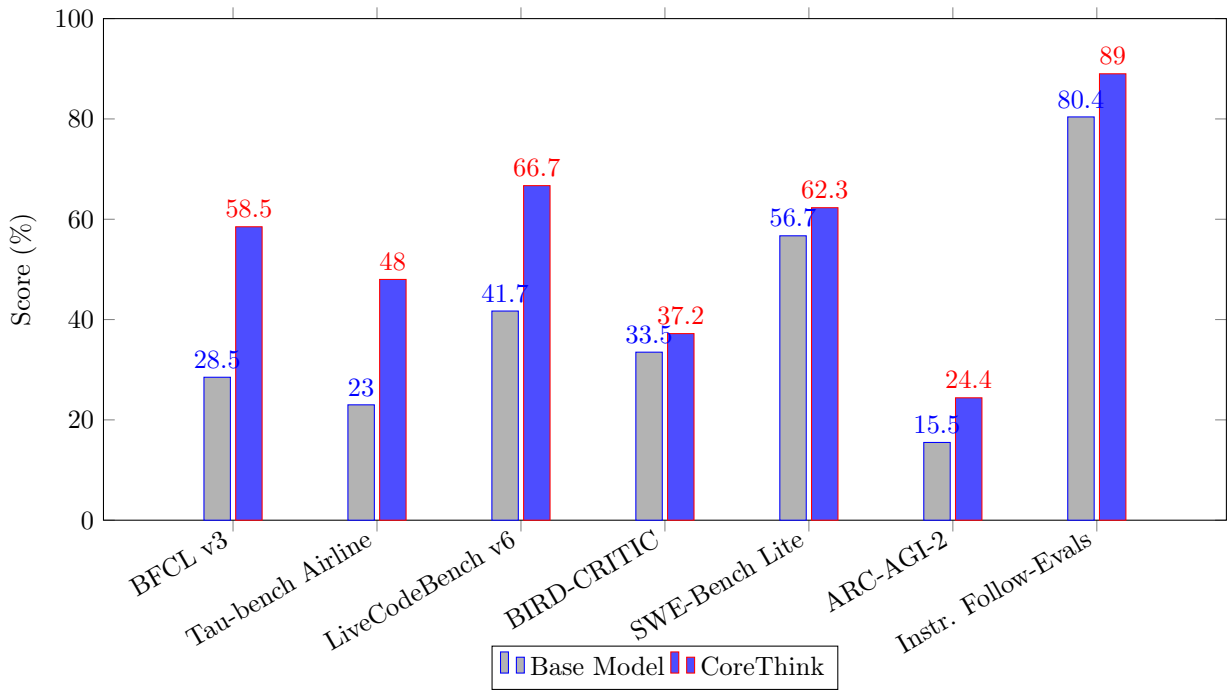
\begin{figure}[htbp]
\centering
\begin{tikzpicture}
\begin{axis}[
    ybar,
    bar width=10pt,
    width=\textwidth,
    height=0.5\textwidth,
    enlarge x limits=0.15,
    ylabel={Score (\%)},
    symbolic x coords={
        BFCL v3,
        Tau-bench Airline,
        LiveCodeBench v6,
        BIRD-CRITIC,
        SWE-Bench Lite,
        ARC-AGI-2,
        Instr. Follow-Evals
    },
    xtick=data,
    x tick label style={rotate=30, anchor=east},
    ymin=0,
    ymax=100,
    legend style={at={(0.5,-0.25)}, anchor=north, legend columns=2},
    nodes near coords,
    nodes near coords align={vertical}
]

\addplot+[style={fill=gray!60}] coordinates {
    (BFCL v3, 28.5)
    (Tau-bench Airline, 23.0)
    (LiveCodeBench v6, 41.7)
    (BIRD-CRITIC, 33.5)
    (SWE-Bench Lite, 56.7)
    (ARC-AGI-2, 15.5)
    (Instr. Follow-Evals, 80.4)
};

\addplot+[style={fill=blue!70}] coordinates {
    (BFCL v3, 58.5)
    (Tau-bench Airline, 48.0)
    (LiveCodeBench v6, 66.7)
    (BIRD-CRITIC, 37.2)
    (SWE-Bench Lite, 62.3)
    (ARC-AGI-2, 24.4)
    (Instr. Follow-Evals, 89.0)
};

\legend{Base Model, CoreThink}
\end{axis}
\end{tikzpicture}
\caption{Side-by-side comparison of CoreThink and baseline model performance across BFCL v3, Taubench-Airline, Livecodebench v6, BIRD-CRITIC, SWE-Bench Lite, ARC-AGI-2, and IF-Evals.}
\label{fig:corethink_vs_base}
\end{figure}

\subsubsection*{Analysis}

\paragraph{Tool-Calling}  
CoreThink outperforms all baselines on \texttt{BFCL v3}, a benchmark that stresses multi-turn interaction and precise function selection. Its symbolic planning layer allows for more coherent multi-step tool invocation. On \texttt{Tau-bench Airline}, where conversational fluidity is emphasized, CoreThink remains competitive though not dominant—highlighting its niche in logical precision over generic dialogic quality.

\paragraph{Code-Generation}  
On \texttt{SWE-Bench Lite}, CoreThink exhibits strong performance due to its capacity for repository-scale reasoning and symbolic planning. This benchmark tests long-horizon comprehension, where standard LLMs tend to falter. While models like \texttt{o4-mini} excel at micro-tasks (e.g., \texttt{BIRD-CRITIC}), CoreThink's strength lies in tasks that demand structural transformation and codebase-level manipulation.

\paragraph{Reasoning and Planning}  
CoreThink's largest margins appear in reasoning tasks such as \texttt{ARC-AGI-2}, which demand abstraction, analogical mapping, and visual logic. With a substantial lead over all baselines, these results suggest the symbolic layer provides new capabilities not present in traditional LLMs. Moreover, its high performance on \texttt{Instruction Following-Evals} demonstrates that this reasoning prowess does not compromise basic command understanding.

\subsubsection*{Synthesis}

These comparative results affirm the impact of the Neuro Symbolic Reasoning layer in elevating the reasoning and decision making capabilities of CoreThink. Future work will expand this study with internal ablations that selectively disable components to further pinpoint their individual contributions.

\section{Conclusion and Future work}
We set out to address the challenges of diminishing returns in large language model performance by introducing CoreThink, a new family of small reasoning models powered by our innovative General Symbolics (GSR) framework. Unlike current dominant approaches, General Symbolics focuses on building a structured, inherently logical reasoning process that directly augments the model, all while operating entirely in natural language. In the future, we will expand this technology to bring GSR to multimodal applications, further enhancing their reasoning capabilities across various data types.

Our evaluations on seven critical benchmarks demonstrate that CoreThink models achieve state-of-the-art performance across their specialized use cases: tool calling, code generation, and reasoning and planning. From an impressive 24.4\% ARC-AGI-2 score on the Berkeley Function Calling Leaderboard to a groundbreaking 62.3\% accuracy on \texttt{SWE-Bench Lite} when integrated into our agentic coding IDE, CoreThink consistently proves its capability and robustness.

We are really excited about what CoreThink and the General Symbolics framework can do for real-world applications. As we continue to refine and expand these models, we invite you to see for yourself how CoreThink can bring truly intelligent and reliable reasoning to your organization.

\section*{Acknowledgments}

The authors would like to thank Ram Shanmugam, Chandra Khatri, and Archie Chaudhury for their helpful discussions and contributions to this work. This research was supported by CoreThink AI.

\printbibliography

\newpage
\appendix
\section{Implementation Details and Development Environment (for Agentic IDE)}
To build our agentic coding IDE, we leveraged key components from the \texttt{SWE-Agent} and \texttt{SWE-Rex} frameworks, adapting their robust environment and tool suite. Our focus was on developing the AI agent itself, integrating the power of our General Symbolics framework. We specifically chose not to build a new connector for \texttt{SWE-Bench Lite}, instead borrowing \texttt{SWE-Agent}'s established connector to ensure seamless and reliable evaluation.

It's worth noting that while we drew inspiration from \texttt{SWE-Agent}, directly embedding General Symbolics into its existing architecture proved challenging. Early trials showed that \texttt{SWE-Agent}'s inherent design was incompatible with the core principles of General Symbolics, leading us to develop the AI agent component independently while still utilizing their excellent environment and tools.

\subsection*{Integrated Tooling for Agentic Operations}
Our agent operates within a rich environment, interacting with the codebase and system through a curated set of tools:
\begin{itemize}
    \item \textbf{Bash Tool:} This is the agent's primary interface for command-line interaction. It allows the execution of virtually any shell command, essential for tasks like running scripts, navigating directories (\texttt{ls -l}), managing Git repositories, and cleaning up temporary files. We configure it to be non-interactive, ensuring consistent and predictable behavior.
    \item \textbf{Tool Bundles:} We utilize several specialized bundles, each providing focused functionalities:
    \begin{itemize}
        \item \textbf{Registry (\texttt{tools/registry}):} Manages the agent's internal state and environment variables. Tools like \texttt{\_read\_env} and \texttt{\_write\_env} allow the agent to store and retrieve information across different steps of a task, maintaining crucial context.
        \item \textbf{Editing (\texttt{tools/edit\_anthropic}):} Provides powerful file modification capabilities. The 
        \\ \texttt{str\_replace\_editor} tool, in particular, enables precise string replacements within files, critical for applying bug fixes or adding new code.
        \item \textbf{Search (\texttt{tools/search}):} Essential for codebase exploration and understanding. Tools like \texttt{find\_file}, \texttt{search\_dir}, and \texttt{search\_file} allow the agent to locate relevant files, understand code connections, and find specific functions or variables (e.g., \texttt{search\_dir "my\_function" "src/"}).
        \item \textbf{Submission (\texttt{tools/review\_on\_submit\_m}):} Contains the \texttt{submit} tool, which the agent uses to finalize its work. This involves generating diffs, presenting changes for review, and providing instructions, ensuring a structured review process.
    \end{itemize}
\end{itemize}

\subsection*{System Architecture and Execution Lifecycle}
The environment's architecture is modular, ensuring a clean, reproducible, and robust development experience for the agent:
\begin{itemize}
    \item \textbf{\texttt{SWEEnv} (The Core Orchestrator):} This central class manages the entire lifecycle of an agent session, from setup to task execution and teardown. It's the primary interface through which our agent interacts with the environment.
    \item \textbf{\texttt{AbstractDeployment} (Deployment Interface):} \texttt{SWEEnv} communicates with the execution backend through this interface, decoupling the agent's logic from the specific environment implementation. Our default and primary choice is \texttt{DockerDeployment}, which provides sandboxed environments for reproducibility.
    \item \textbf{\texttt{Repo} (Codebase Manager):} Responsible for maintaining the state of the software repository. Before each task, it ensures the codebase is consistent by cloning the repository and checking out the specified base commit.
    \item \textbf{\texttt{CombinedEnvHooks} (Extensibility Mechanism):} A hook system allows us to inject custom functionalities at various points in the environment's lifecycle, useful for logging, monitoring, or adding custom instrumentation.
\end{itemize}
The environment follows a strict execution lifecycle to guarantee clean and reproducible agent sessions:
\begin{itemize}
    \item \textbf{Initialization:} The environment is set up with a declarative configuration, defining the Docker image and target repository.
    \item \textbf{Setup (\texttt{start}):} The environment launches the Docker container, sets its initial state, and executes any pre-start commands.
    \item \textbf{Task Reset (\texttt{reset}):} Crucially, before each new task or retry, the \texttt{reset} method restores the environment to its pristine state. This involves resetting the file system and version control to the specified base commit, ensuring every agent attempt starts from identical conditions.
    \item \textbf{Termination (\texttt{close}):} Upon task completion or failure, the \texttt{close} method gracefully terminates the deployment, stopping the container and releasing resources.
\end{itemize}

\subsection*{Agent-Environment Interface (API)}
The \texttt{SWEEnv} class exposes a well-defined API, enabling the agent to perceive and act on its environment:
\begin{itemize}
    \item \textbf{Command Execution:}
    \begin{itemize}
        \item \texttt{communicate(command: str) -> str:} The primary method for agent interaction, executing a command within a stateful shell session and returning its standard output.
        \item \texttt{execute\_command(...):} Executes a command in a new, non-interactive process, suitable for background tasks.
    \end{itemize}
    \item \textbf{File System Access:}
    \begin{itemize}
        \item \texttt{read\_file(path: str) -> str:} Retrieves the contents of a file.
        \item \texttt{write\_file(path: str, content: str):} Writes or overwrites a file.
    \end{itemize}
    \item \textbf{State Management:}
    \begin{itemize}
        \item \texttt{set\_env\_variables(vars: dict):} Sets environment variables within the active shell session.
    \end{itemize}
\end{itemize}
This comprehensive setup provides the robust and controlled environment necessary for CoreThink's agentic capabilities, enabling it to effectively tackle complex software engineering challenges.

\newpage
\section{System Details for ARC-AGI-2 Neuro-Symbolic Pipeline}
\addcontentsline{toc}{section}{Appendix B: ARC-AGI-2 Neuro-Symbolic Pipeline}

To address the unique challenges posed by the ARC-AGI-2 benchmark, we developed a four-stage neuro-symbolic framework called \texttt{ARC-AGI NS Flow}. This hybrid pipeline fuses deterministic grid-level perception with symbolic rule synthesis and neural language model execution, enabling high generalization performance in few-shot abstract reasoning tasks.

\subsection*{Stage 1: Deterministic Object Detection}
We begin with a rule-based traversal of the input grid to identify discrete visual objects:
\begin{itemize}
    \item \textbf{BFS-based segmentation:} We traverse all connected components using a color-aware breadth-first search.
    \item \textbf{Object masks:} Each object's pixel set is encoded for downstream analysis.
    \item \textbf{Cavity analysis:} Topological features (e.g., holes) are recorded per object.
    \item \textbf{Background profiling:} Dominant background color is tracked to distinguish figure-ground structure.
\end{itemize}
This stage runs in $\mathcal{O}(N)$ time and is deterministic, providing a reproducible perception layer for all inputs.

\subsection*{Stage 2: Unit Pattern Detection via LLM}
To extract symbolic abstractions, we define a taxonomy of 23 atomic operations (e.g., \texttt{translate}, \texttt{reflect}, \texttt{cavity\_fill}) grounded in object-level transformations:
\begin{itemize}
    \item \textbf{Change tagging:} Objects are labeled as \texttt{added}, \texttt{removed}, or \texttt{retained} between each input-output training pair.
    \item \textbf{Pattern classification:} We prompt \texttt{o4-mini} with object metadata to classify transformation patterns and extract parameters.
    \item \textbf{Symbolic abstraction:} Detected transformations are abstracted as symbolic units for downstream intersection.
\end{itemize}
This step compresses visual changes into interpretable rule tokens.

\subsection*{Stage 3: Pattern Intersection and Constraint Synthesis}
Symbolic detections across training examples are intersected to derive a unified transformation rule:
\begin{itemize}
    \item \textbf{Aggregation:} All detected unit patterns are collected and ranked by frequency/confidence.
    \item \textbf{Filtering:} Low-confidence or inconsistent patterns are discarded via self-consistency heuristics.
    \item \textbf{Constraint generation:} Surviving patterns are translated into symbolic hints (e.g., 'rotate the red cavity 90 ° clockwise').
\end{itemize}
The resulting rule set captures generalizable transformations and contextual logic.

\subsection*{Stage 4: Solving via Self-Consistent LLM Prompting}
We use \texttt{Grok-4} as our final symbolic executor, guided by multi-shot prompting:
\begin{itemize}
    \item \textbf{Prompt format:} Grids are encoded in markdown-style tables, alongside training demonstrations and symbolic hints.
    \item \textbf{Self-consistency voting:} The model generates multiple samples; each test output is resolved by majority vote per pixel.
    \item \textbf{Fallback strategy:} In 2-pass settings, we back off to base \texttt{Grok-4} when our pipeline fails on attempt 1.
\end{itemize}
This method balances symbolic precision with neural flexibility.

\subsection*{Performance Summary}
Our system achieves 24.4\% accuracy on the full ARC-AGI-2 evaluation set, outperforming all known symbolic, neural, and hybrid baselines by a significant margin. The top LLM-only baseline (Grok-4) trails by over 8 points, highlighting the advantage of neuro-symbolic reasoning.

\subsection*{Challenges and Engineering Notes}
\begin{itemize}
    \item \textbf{Pixel sensitivity:} Grid outputs must be exact, making robustness to LLM misalignment or off-by-one errors critical.
    \item \textbf{Resolution heuristics:} Downscaling grids improved accuracy but limited generalization; future work will adopt adaptive encodings.
    \item \textbf{Symbolic noise:} Intersected pattern constraints sometimes include spurious rules. Pruning and soft constraints are under exploration.
    \item \textbf{Coord encoding:} LLMs occasionally misinterpret grid formats; structured encodings and token-level coordinate schemes are being developed.
\end{itemize}

This appendix supplements our main ARC-AGI section by outlining system architecture, reasoning stages, and practical constraints encountered during ARC-AGI-2 development. We hope this pipeline can serve as a stepping stone for future neuro-symbolic AGI work.

\end{document}